\documentclass[twoside,leqno,twocolumn]{article}
\usepackage{ltexpprt}
\usepackage{graphicx}
\usepackage{amsmath}
\usepackage{algorithmic}
\usepackage{algorithm}
\usepackage{amssymb}
\usepackage{subfigure}
\usepackage{epsfig}
\usepackage{amsfonts}
\usepackage{times}
\usepackage{yhmath}
\usepackage{footnpag}
\usepackage{multicol}
\usepackage{multirow}
\usepackage{psfrag}
\usepackage{verbatim}
\usepackage{cite}
\usepackage[utf8]{inputenc}
\usepackage{float}
\begin{document}

\title{\Large  Towards Scalable Spectral Clustering via Spectrum-Preserving Sparsification
}
\author{Yongyu Wang\thanks{Michigan Technological University, Houghton, MI, 49931, USA. E-mail: yongyuw@mtu.edu.} \\
\and
Zhuo Feng\thanks{Michigan Technological University, Houghton, MI, 49931, USA. E-mail: zhuofeng@mtu.edu.}}
\date{}

\maketitle


\fancyfoot[R]{\scriptsize{Copyright \textcopyright\ 20XX by SIAM\\
Unauthorized reproduction of this article is prohibited}}





\begin{abstract} \small\baselineskip=9pt Eigenvalue decomposition of Laplacian matrices for large nearest-neighbor (NN) graphs is the major computational bottleneck in spectral clustering (SC). To fundamentally address this computational challenge in SC, we propose a  scalable spectral sparsification framework that enables to construct nearly-linear-sized ultra-sparse NN graphs with guaranteed preservation of key eigenvalues and eigenvectors of the original Laplacian. The proposed method is based on the latest theoretical results in spectral graph theory and thus can be  applied to robustly handle general undirected graphs. By leveraging a nearly-linear time spectral graph topology sparsification phase and a subgraph scaling phase via stochastic gradient descent (SGD) iterations, our approach allows computing tree-like NN graphs that can serve as high-quality  proxies of the original NN graphs,  leading to highly-scalable and accurate   SC of large data sets.  Our extensive experimental results on a variety of public domain data sets show  dramatically improved clustering performance when compared with state-of-the-art  SC methods. For instance, SC of a large  data set \textbf{Covtype} that has $581,012$ instances using our method only takes $9.43$ seconds (compared to $91,504$ seconds with the original one) without loss of clustering accuracy.\end{abstract}


\section{Introduction}
Data clustering and graph partitioning  are playing increasingly important roles in  many numerical and graph related applications such as scientific computing, data mining, machine learning,  image processing, etc. Among the existing data clustering and graph partitioning techniques, spectral methods have gained great attention in recent years \cite{spielmat1996spectral,ng2002spectral,kolev2015note,peng2015partitioning}, which typically require to solve eigenvalue decomposition problems associated with  graph Laplacians. For example, classical spectral clustering (SC) or partitioning algorithms leverage the  first few  graph Laplacian eigenvectors corresponding to the smallest  nontrivial (nonzero) eigenvalues for low dimensional graph (data points) embedding, which is followed by a k-means clustering procedure that usually leads to high-quality   clustering (partitioning) results. Although spectral methods  have many advantages, such as easy implementation, good clustering/partitioning quality and rigorous theoretical foundations \cite{lee2014multiway,kolev2015note,peng2015partitioning}, the high computational (memory and runtime) cost  due to the involved eigenvalue decomposition procedure can immediately hinder their applications in emerging big data (graph) analytics tasks \cite{chen2011parallel}.

To address the computational bottleneck of   spectral methods for data clustering or graph partitioning, recent research efforts  aim to   reduce the complexity of the original data network (graph Laplacian) through various kinds of approximations: k-nearest neighbor (kNN) graphs  maintain  k nearest neighbors (NNs) for each node (data point), whereas $\epsilon$-neighborhood graphs  keep the neighbors within the range of distance $\epsilon$ \cite{muja2014scalable};  a sampling-based approach for affinity matrix approximation using  Nystr\"om method has been introduced in \cite{fowlkes2004spectral,williams2001using}, while its error analysis has been introduced in \cite{choromanska2013fast,zhang2008improved};  a landmark-based method for representing the original data points has been  introduced in \cite{chen2011large}; \cite{yan2009fast} proposed a general framework for fast approximate SC by collapsing the original data points into a small number of centroids using k-means and random-projection trees; \cite{liu2013large} introduced a method for compressing the original NN graph into a sparse bipartite graph by generating a small number of ``supernodes"; a graph sparsification method using a simple similarity-based heuristic has been proposed for scalable clustering \cite{satuluri2011local}; \cite{li2016scalable} attempted to perform clustering sequentially. However, none of these approximation methods can reliably preserve the spectrums of the original graphs, and thus may lead to degraded clustering  result. For example, spectral clustering of the \textbf{Covtype} data set using the existing approximation methods lead to dramatically degraded solution quality: the clustering accuracy  drops from $48.83\%$ to $22.80\%$ when using the landmark based approach \cite{chen2011large}. Other approximation methods also lead to more than  $20\%$ loss of accuracy. 

Recent {{spectral graph sparsification  }} research  results from theoretical computer science (TCS) community show the possibility of computing nearly-linear-sized\footnote{ The number of edges is similar to the number of nodes in the sparsifier.} graph sparsifiers (proxies) for preserving the spectrum (the first few eigenvalues and eigenvectors) of the original graph (Laplacian), which can potentially lead to  a series of {\emph{``theoretically nearly-linear"}}  numerical and graph algorithms for solving sparse  matrices, graph-based semi-supervised learning (SSL), as well as spectral graph (data) partitioning (clustering) and max-flow problems \cite{Spielman:usg,spielman2011graph,spielman2010algorithms,Kolla:2010stoc,miller:2010focs, Fung:2011stoc,spielman2011spectral, christiano2011flow, spielman2014sdd}.  For instance,   sparsified social networks allow to more effectively understand and predict information propagation phenomenons in large social networks;  sparsified data networks allow  more efficiently storing, partitioning (clustering) and analyzing big data networks; sparsified matrices can be leveraged to accelerate solving large linear system of equations.

Inspired by the latest progress in spectral graph sparsification research \cite{feng2016spectral,spielman2011spectral,batson2013spectral, feng2018similarity, Lee:2017}, we propose a highly-scalable two-phase spectral graph sparsification framework  for extracting nearly-linear-sized ultra-sparse NN graphs from the original NN graphs. In contrast with the recent sampling-based spectral sparsification method  \cite{chen2016communication} that is only suitable for handling extremely dense kNN graphs (e.g. k=5,000), our method is more general and can  deal with both dense and sparse kNN graphs while producing ultra-sparse spectrally-similar sparsifiers. Our proposed framework includes a graph topology sparsification phase and an edge weight scaling phase, which can immediately leads to  highly efficient and accurate SC of large data sets. The key contributions of this work have been summarized as follows:
\begin{enumerate}
  \item Compared with existing approximation approaches that have no guarantee of the solution quality in SC, the proposed spectrum-preserving  sparsification framework can robustly preserve the most critical spectral properties of the original graph, such as the first few eigenvalues and eigenvectors of graph Laplacians, within \textbf{nearly-linear-sized tree-like subgraphs} that allows for computing high-quality clustering results.
  \item A practically-efficient spectral graph topology sparsification method \cite{feng2016spectral,feng2018similarity} has been applied for  identifying spectrally critical edges to be kept in subgraphs, while a novel \textbf{scalable subgraph scaling scheme} via stochastic gradient descent (SGD) iterations has been proposed to scale up edge weights to further improve the spectral approximation of the subgraph.
  \item Motivated by recent graph signal processing research\cite{shuman2013emerging,sandryhaila2014big}, we show that spectral graph sparsification can be considered as a \textbf{``low-pass" graph filter} for removing edges less critical for preserving the first few graph Laplacian eigenvectors. We also introduce a simple yet effective procedure for filtering out errors in Laplacian eigenvectors computed using spectrally sparsified graphs, which  enables to  leverage much sparser subgraphs in SC   for achieving superior solution quality. 
  
  \item We have conducted  extensive experiments using the proposed method and compare against prior state-of-the-art  methods  for several well-known public domain data sets. Our experiment results show that the proposed method can dramatically improve SC efficiency  as well as accuracy. For instance, SC based on the spectrally sparsified \textbf{Covtype}  graph  only takes less than two minutes (compared to many hours based on the original graph), while producing even higher clustering accuracy ($48.9\%$ compared to $\%44.2$).
\end{enumerate}


\section{Preliminaries}\label{sect:preliminaries}
\subsection{Spectral Clustering Algorithms}
SC algorithms can often outperform traditional clustering algorithms, such as k-means algorithms \cite{von2007tutorial}. 
Consider a similarity graph $G=(V,E_G,\mathbf{w_G})$, where $V$ and $E_G$ denote the graph vertex and edge sets, respectively, while  $\mathbf{w}_G$ denotes a weight (similarity) function that assigns positive weights to all edges. The  symmetric diagonally dominant (SDD) Laplacian matrix of graph  $G$  can be constructed as follows:
\begin{equation}\label{formula_laplacian}
\mathbf{L}_G(i,j)=\begin{cases}
-w_{ij} & \text{ if } (i,j)\in E_G \\
\sum\limits_{(i,k)\in E_G} w_{ik} & \text{ if } (i=j) \\
0 & \text{otherwise }.
\end{cases}
\end{equation}
SC algorithms typically include the following three steps: 1) construct a Laplacian matrix according to the similarities between data points; 2) embed nodes (data points) into $k$-dimensional space using the first $k$ nontrivial graph Laplacian eigenvectors; 3) apply k-means algorithm to partition the embedded data points into $k$ clusters. The standard SC algorithms \cite{von2007tutorial} can be very computationally expensive for  handling large-scale data sets due to the very costly procedure for computing Laplacian eigenvectors, whereas prior approximate SC algorithms \cite{fowlkes2004spectral,yan2009fast,chen2011large,liu2013large}  can not always guarantee solution quality.

 \subsection{Spectral Graph Sparsification}
 Graph sparsification aims to find a subgraph (sparsifier) $S=(V,E_S,\mathbf{w}_S)$ that has the same set of vertices of the original graph $G=(V,E_G,\mathbf{w}_G)$, but  much fewer edges. We say  $G$ and its subgraph $S$ are { $\sigma$-spectrally similar} if the following condition holds for all real vectors $\mathbf{x} \in \mathbb{R}^V$ 
\begin{equation}
\label{formula_spectral_similar}
\frac{\mathbf{x}^\top{\mathbf{L}_{S}}\mathbf{x}}{\sigma}\le \mathbf{x}^\top{\mathbf{L}_G}\mathbf{x} \le \sigma \mathbf{x}^\top{\mathbf{L}_{S}}\mathbf{x},
\end{equation}
where $\mathbf{L}_{G}$ and $\mathbf{L}_{S}$ denote the symmetric diagonally dominant (SDD) Laplacian matrices  of  $G$ and $P$, respectively.
By defining the relative condition number to be $\kappa({\mathbf{L}_G},{\mathbf{L}_{S}})=\lambda_{\max}/\lambda_{\min}$, where $\lambda_{\max}$ ($\lambda_{\min}$) denotes the largest (smallest  nonzero) eigenvalues of $\mathbf{L_S^{+} L_G}$, and $\mathbf{L_S^+}$ denotes the Moore-Penrose pseudoinverse of $\mathbf{L_S}$,  it can be further shown that $\kappa(\mathbf{L_G},\mathbf{L_S})\le\sigma^2$, indicating that a smaller relative condition number or $\sigma^2$ corresponds to a higher (better) spectral similarity.

\section{SGD-based Two-Phase Spectral Graph Sparsification}\label{sect:two_phase}
In the following, we assume that $G=(V,E_G,\mathbf{w_G})$ is a weighted, undirected, and connected graph, whereas $S=(V,E_S,\mathbf{w}_S)$ is its graph sparsifier. The descending   eigenvalues of $\mathbf{L}^+_{S} \mathbf{L}_G$ are denoted by $\lambda_{\max}={\lambda _1} \ge {\lambda _2} \ge \cdots \ge {\lambda _n} \ge 0$, where $\mathbf{L}^+_{S}$ denotes the Moore-Penrose pseudoinverse of $\mathbf{L}_{S}$.

\subsection{Phase 1: Spectral Graph Topology Sparsification}\label{sec:top_spar}
\subsubsection{Off-Tree Edge Embedding with Approximate Generalized Eigenvectors}
Spielman~\cite{spielman2009note} showed that $\mathbf{L}^+_{S} \mathbf{L}_G$ has at most $k$   eigenvalues greater than $\textstyle{{\rm{st}}}_{S}(G)/k$, where  $\textstyle{{\rm{st}}}_{S}(G)$ is the total stretch of the spanning-tree subgraph $S$ with respect to the original graph $G$. The total stretch can be considered as the spectral distortion due to the subgraph approximation. Recent results show that every graph has a { low-stretch spanning tree} (LSST) with bounded total stretch \cite{abraham2012using}:
 \begin{equation}\label{formula_stretch}
O(m \log m \log \log n)\geq \textstyle{{\rm{st}}}_{S}(G) =\\ \textstyle{{\rm{tr}}}({\mathbf{L}^+_{S} \mathbf{L}_G})\geq  \kappa({\mathbf{L}_G},{\mathbf{L}_{S}}),
\end{equation}
where $m=|E_G|$, $n=|V|$, and $\textstyle{{\rm{tr}}}({\mathbf{L}^+_{S} \mathbf{L}_G})$ is the trace of ${\mathbf{L}^+_{S} \mathbf{L}_G}$. Such a theoretical result allows constructing $\sigma$-similar spectral sparsifiers with $O(\frac{m \log n \log \log n}{\sigma^2})$ off-tree edges   in nearly-linear time using a spectral perturbation framework \cite{feng2016spectral,feng2018similarity}. To identify the key off-tree edges to  be added to the low-stretch  spanning tree (LSST) for dramatically reducing spectral distortion (the total stretch),   a spectral embedding scheme using approximate generalized eigenvectors has been introduced in \cite{feng2016spectral}, which is based on the following spectral perturbation analysis:
\begin{equation}\label{formula_eig_perturb1}
\mathbf{L}_G({\mathbf{u}_i + \delta \mathbf{u}_i}) = ({\lambda _i + \delta \lambda _i})(\mathbf{L}_{S} + \delta \mathbf{L}_{S})({\mathbf{u}_i + \delta \mathbf{u}_i}),
\end{equation}
where a perturbation $\delta \mathbf{L}_{S}$ is applied to $\mathbf{L}_{S}$, which results in perturbations in generalized eigenvalues ${\lambda _i} + \delta {\lambda _i}$ and eigenvectors ${\mathbf{u}_i} + \delta {\mathbf{u}_i}$ for $i=1, \ldots, n$, respectively. The first-order perturbation analysis \cite{feng2016spectral} leads to
\begin{equation}\label{formula_eig_perturb5}
-\frac{\delta {\lambda _i}}{{\lambda _i}} = \mathbf{u}_i^\top\delta \mathbf{L}_{S}{\mathbf{u}_i},
\end{equation}
which indicates that the reduction of $\lambda _i$ is proportional to the Laplacian quadratic form of $\delta \mathbf{L}_{S}$ with the generalized eigenvector $\mathbf{u}_i$. (\ref{formula_eig_perturb5}) can also be understood through the following \textbf{Courant-Fischer theorem } for generalized eigenvalue problems:
\begin{equation}\label{formula_courant-fischer-max}
\lambda_{1}=\mathop{\max_{|\mathbf{x}|\neq 0}}_{\mathbf{x}^\top\mathbf{1}=0}\frac{\mathbf{x}^\top \mathbf{L}_G \mathbf{x}}{\mathbf{x}^\top \mathbf{L}_S \mathbf{x}}\approx\mathop{\max_{|\mathbf{x}|\neq 0}}_{x(p)\in \left\{0,1 \right\}}\frac{\mathbf{x}^\top \mathbf{L}_G \mathbf{x}}{\mathbf{x}^\top \mathbf{L}_S \mathbf{x}}=\mathop{\max}\frac{|\partial_G (Q)|}{|\partial_S (Q)|},
\end{equation}
where  $\mathbf{1}$ is the all-one vector, the node set $Q$ is defined as $Q \overset{\mathrm{def}}{=}\left\{ p \in V: x(p)=1 \right\}$, and the boundary of $Q$ in $G$ is defined as
$\partial_G (Q) \overset{\mathrm{def}}{=}\left\{ (p,q)\in E_G: p\in Q, q\notin Q\right\}$, which obviously leads to $
\mathbf{x}^\top \mathbf{L}_G \mathbf{x}=|\partial_G (Q)|,\mathbf{x}^\top \mathbf{L}_S \mathbf{x}=|\partial_S (Q)|$.

 Then (\ref{formula_courant-fischer-max}) indicates that  finding the dominant generalized eigenvector is approximately equivalent to problem of finding $Q$ such that $\frac{|\partial_G (Q)|}{|\partial_S(Q)|}$ or the mismatch of boundary (cut) size between the original graph $G$ and subgraph $S$ is maximized. As a result, $\lambda_{max}=\lambda_{1}$ is a good measure of the largest  mismatch in boundary (cut) size between $G$ and $S$. Once $Q$ or $\partial_S(Q)$ is found via graph embedding using dominant generalized eigenvectors, we can selectively pick the   edges from $\partial_G (Q)$ and  recover them to $S$ to dramatically  mitigate maximum mismatch or $\lambda_{1}$. 

Denote $\mathbf{e}_{p}\in \mathbb{R}^V$ the vector with only the $p$-th element being $1$ and others being $0$. We also denote $\mathbf{e}_{pq}=\mathbf{e}_{p}-\mathbf{e}_{q}$, then the generalized eigenvalue perturbation due to the inclusion of off-tree edges can be expressed as follows
\begin{equation}\label{formula_eig_perturb6}
-\frac{\delta {\lambda _i}}{{\lambda _i}}=\mathbf{u}_i^\top \delta \mathbf{L}_{S,max}\mathbf{u}_i = \sum_{(p,q)\in E_G\setminus E_S}^{}{{w_{pq}}\left(\mathbf{e}_{pq}^T\mathbf{u}_i\right)^2},
\end{equation}
where $\delta\mathbf{L}_{S,max}=\mathbf{L}_{G}-\mathbf{L}_{S}$. The \textbf{spectral criticality} $c_{pq}$ of each off-tree edge $(p,q)$ is defined as:
 \begin{equation}\label{formula_criticality}
c_{pq}={{w_{pq}}\left(\mathbf{e}_{pq}^T\mathbf{u}_1\right)^2}\approx {{w_{pq}}\left(\mathbf{e}_{pq}^T\mathbf{h}_t\right)^2},~~~~ {\mathbf{h}_t}= \left({\mathbf{L}_{S}^{+}}\mathbf{L}_G\right)^t{\mathbf{h}_0}, 
\end{equation}
where $\mathbf{h}_t$ denotes the approximate dominant generalized eigenvector computed through a small number (e.g. $t=2$) of generalized power iterations using an initial random vector $\mathbf{h}_0$. (\ref{formula_criticality}) can be regarded as the edge Joule heat (power dissipation) when considering the undirected graph $G$ as a resistor network and $\mathbf{u}_1$ or $\mathbf{h}_t$ as the voltage vector, which can  be leveraged for identifying and recovering the most spectrally critical off-tree edges from $\partial_G(Q)$ into LSST  for spectral graph topology sparsification,  which allows dramatically reducing largest mismatch ($\lambda_{1}$) and thus improving spectral approximation in the subgraph.

 \subsubsection{A Trace Minimization Perspective for Spectral Sparsification. }
It is critical to assure that recovering the  most ``spectrally critical" off-tree edges identified by (\ref{formula_criticality})  can always effectively improve the preservation of  key (bottom) eigenvalues and eigenvectors within the  sparsified Laplacians. For example, in spectral clustering tasks, the bottom few eigenvectors of (normalized) data graph Laplacian matrix are used for embedding high-dimensional data points onto low-dimensional ones such that the following k-means procedure can effectively group data points into high-quality clusters \cite{shi2000normalized,ng2002spectral}. Consequently, robustly preserving the graph spectrum in the sparsified graph will be key to scalable yet accurate spectral graph (data) analysis.

Consider the following analysis for undirected graphs. Denote the descending  eigenvalues and  the corresponding unit-length, mutually-orthogonal eigenvectors  of $\mathbf{L}_G$  by ${\zeta_1} \geq \cdots > {\zeta _n}=0$, and ${\omega _1}, \cdots, {\omega _n}$, respectively. Similarly denote the eigenvalues and eigenvectors  of $\mathbf{L}_S$ by ${\tilde \zeta _1} \geq \cdots > {\tilde \zeta _n}=0$ and ${\tilde \omega _1}, \cdots, {\tilde \omega _n}$, respectively. It should be noted that both $\omega_n$ and $\tilde \omega _n$ are the normalized and orthogonal to the all-one vector $\mathbf{1}/\sqrt[]{n}$. Then the following spectral decompositions of  $\mathbf{L}_G$ and $\mathbf{L}^+_S$ always hold:

\begin{equation}\label{formula_decomposition G}
\mathbf{L}_G = \sum\limits_{i = 1}^{n-1} {{\zeta _i}{\omega_i}{\omega^\top_i}},~~~~~~~~ \mathbf{L}_S^{+} = \sum\limits_{j = 1}^{n-1} {{{\frac{1}{\tilde \zeta_j}}}{\tilde \omega_j}{\tilde \omega^\top _j}},
\end{equation}
which leads to the following trace of $\mathbf{L}_S^{+}\mathbf{L}_G$:
\begin{equation}\label{formula_combine trace 1}
\textbf{Tr}({\mathbf{L}_S^{+}}\mathbf{L}_G) ={\textbf{Tr}}(\sum\limits_{i = 1}^{n-1} \sum\limits_{j = 1}^{n-1} {{\frac{\zeta _i}{\tilde \zeta_j}}{\epsilon_{ij}}{\tilde \omega_j}{\omega^\top _i}})=\sum\limits_{j = 1}^{n-1} \frac{1}{\tilde \zeta_j} \sum\limits_{i = 1}^{n-1} {{\zeta _i}}{\epsilon^2_{ij}},
\end{equation}
where $\epsilon_{ij}$ satisfies: $0 \leq \epsilon^2_{ij} = ({\omega^\top _i}{\tilde \omega_j})^2\leq 1$.
According to  (\ref{formula_combine trace 1}), the most spectrally critical off-tree edges identified by (\ref{formula_criticality}) will impact the largest eigenvalues of $\mathbf{L}_S^{+}\mathbf{L}_G$ as well as the bottom (smallest nonzero) eigenvalues of $\mathbf{L}_S$, since the smallest $\tilde \zeta_j$ values directly contribute to the largest components in the trace of $\mathbf{L}_S^{+}\mathbf{L}_G$. This fact   enables to  recover small portions  of most spectrally critical off-tree edges to LSST subgraph for preserving the key spectral graph properties within the sparsified graph.

 \subsubsection{A Scheme for Eigenvalue Stability Checking.}
\begin{figure}
\includegraphics[scale=0.44]{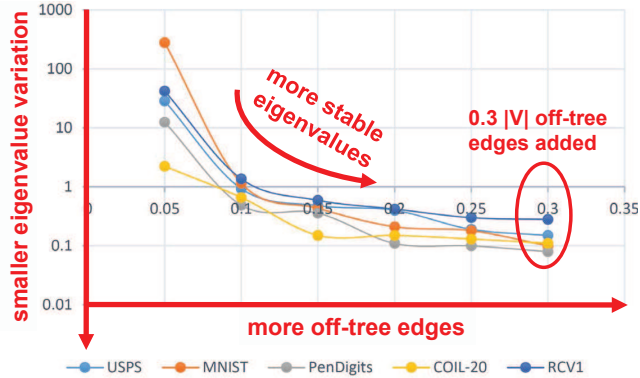}
\caption{Variation ratio of bottom eigenvalues with increasing  number of off-tree edges.\protect\label{fig:eig_stable}}
\end{figure}

We propose a novel method for checking the stability of bottom eigenvalues of the sparsified Laplacian. Our approach proceeds as follows: 1) in each iteration for recovering off-tree edges, we compute and record the several smallest eigenvalues of the latest sparsified Laplacian: for example, the bottom k eigenvalues that are critical for spectral clustering tasks; 2) we determine whether more off-tree edges should be recovered by looking at the stability by comparing with the eigenvalues computed in the previous iteration: if the change of eigenvalues  is significant, more off-tree edges should be added to the current sparsifier. More specifically, we store the bottom k eigenvalues computed in the previous (current) iteration  into vector $v_p$ ($v_{p+1}$), and calculate the eigenvalue variation ratio  by:
 \begin{equation}\label{eqn:varaition ratio}
ratio_{var} = \frac{{\|(v_p-v_{p+1})\|}}{\|(v_p)\|}.
\end{equation}
A greater eigenvalue variation ratio indicates less stable eigenvalues within the latest sparsified graph Laplacian, and thus justifies another iteration to allow adding more "spectrally-critical" off-tree edges into the sparsifier. As shown in Fig. \ref{fig:eig_stable}, for a variety of kNN graphs obtained from public domain data sets \cite{chen2011parallel,li2016scalable}, adding a small portion of off-tree edges will immediately stabilize the distributions of the first few (normalized) Laplacian eigenvalues. 
\subsubsection{Inexact Yet Fast Eigenvalue  Computation.}
Among various eigenvalue solvers, the software package  ARPACK has become the standard solver for solving practical large-scale eigenvalue problems \cite{lehoucq1997arpack}. The MATLAB also uses ARPACK as the eigensolver for sparse matrix problems. \cite{chen2011parallel} shows that the ARPACK employs implicitly iterative restarted Arnoldi  process that contains at most ($z-k$) steps, where $z$ is the Arnoldi length empirically set to $2k$ for sparse matrices. Since the cost of each iteration is $O(nw)$ for a sparse matrix-vector product, the overall runtime cost of ARPACK solver is proportional to  $O(z^3)+(O(nz)+O(nw))\times O(z-k)$, and the memory cost is $O(nw)+O(nz)$, where $n$ is the number of data points, $z$ is the arnoldi length, $w$ is the number of nearest neighbors, and $k$ is the number of desired eigenvalues.

Algorithms with a sparsified Laplacian can dramatically reduce the time and memory cost of the ARPACK solver due to the dramatically reduced $w$. To gain high efficiency,  we propose to quickly compute  eigenvalues for stability checking based on an inexact implicitly restarted Arnoldi method \cite{xue2009numerical}. It has been shown  that by relaxing the tolerance, the total inner iteration counts can be significantly reduced, while the inner iteration cost can be further reduced by using subspace recycling with iterative linear solver.

\subsection{Phase 2: Subgraph Scaling via SGD Iterations}
 To aggressively limit the number of edges in the subgraph $S$ while still achieving a high quality approximation of the original graph $G$, we introduce an efficient edge scaling scheme to mitigate the accuracy loss. We propose to scale up edge weights in the subgraph $S$ to further reduce the largest mismatch or $\lambda_1$. The dominant eigenvalue perturbation $\delta \lambda_1$ in terms of edge weight perturbations using first-order analysis (\ref{formula_eig_perturb1}) can be expressed as: 
\begin{equation}\label{formula_weight_sensitivity}
-\frac{\delta {\lambda _1}}{{\lambda _1}}= \mathbf{u}^\top_1\delta \mathbf{L}_{S}\mathbf{u}_1 = \sum_{({p,q})\in E_S}^{} {{\delta w_{pq}}\left(\mathbf{e}_{pq}^\top\mathbf{u}_1\right)^2},
\end{equation}

\noindent which directly gives the sensitivity of $\lambda _1$ with respect to each edge weight $w_{pq}$ as follows:

 \begin{equation}\label{formula_weight_sensitivity2}
  \frac{\delta \lambda _1}{\delta w_{pq}}=-\lambda _1 \left(\mathbf{e}_{pq}^\top\mathbf{u}_1\right)^2\approx -\lambda _1 \left(\mathbf{e}_{pq}^\top\mathbf{h}_t\right)^2.
\end{equation}
 
 \noindent With the (approximate) sensitivity expressed in (\ref{formula_weight_sensitivity2}), it is possible to find a proper weight scaling factor for each edge in $S$ such that $\lambda_1$ will be dramatically reduced. However, since both $\lambda_1$ and $\lambda_n$ will decrease monotonically when scaling up edge weights, it is  likely that $\lambda_n$ will decrease at a faster rate than $\lambda_1$, which will lead to even a worse spectral approximation. To address this issue, we adopt nearly-linear time algorithms for estimating the extreme generalized eigenvalues $\lambda_1$ and $\lambda_n$ introduced in \cite{feng2018similarity}, which allows us to scale edge weight properly without degrading the spectral approximation quality.
 
Then the following constrained nonlinear optimization framework can be leveraged for scaling up the subgraph ($S$) edge weights $\mathbf{w}_s$ to minimize the largest mismatch reflected by the largest generalized eigenvalue $\lambda_1$ in (\ref{formula_courant-fischer-max}).
\begin{equation}\label{edge_opt}
\begin{split}
  {~~~\textbf{minimize:}} ~~~~~ \lambda_{1}(\mathbf{w}_s)\\
 {\textbf{subject to:}}~~~~~~~~~~~~~~~~\\
\textbf{(a)}~~~~~~~ \mathbf{L}_{G} \mathbf{u}_i&=\lambda_{i} \mathbf{L}_{S}\mathbf{u}_i,~~i=1,...,n;~~~~~~\\
\textbf{(b)}~~~~~~~\lambda^{}_{max}&=\lambda^{}_{1}\geq \lambda^{}_{2}...\geq \lambda^{}_{n}=\lambda^{}_{min};\\
\textbf{(c)}~~~~~~~~\lambda^{}_{n}~~~~&\geq \lambda^{(0)}_{n}\overline{\Delta}_{\lambda_{n}}.
\end{split}
\end{equation}

In (\ref{edge_opt}), $\lambda^{(0)}_{n}$ and $\lambda_{n}$ denote the smallest nonzero eigenvalues before and after  edge scaling, respectively, whereas $\overline{\Delta}_{\lambda_{n}}$ denotes the upper bound of reduction factor in  $\lambda^{(0)}_{n}$ after edge scaling. (\ref{edge_opt}) aims to minimize $\lambda_{1}$ by scaling up subgraph edge weights  while limiting the decrease in $\lambda_{n}$.

\begin{algorithm}[!htbp]
\small { \caption{Subgraph Edge Weight Scaling via Constrained SGD Iterations} \label{alg:sgd}
\textbf{Input:} $\mathbf{L}_G$, $\mathbf{L}_S$, $\mathbf{d}_G$, $\mathbf{d}_S$, $\lambda_1^{(0)}$, $\lambda_n^{(0)}$, $\overline{\Delta}_{\lambda_{n}}$, $\beta$, $\eta_{max}$, $\epsilon$, and $N_{max}$\\
\textbf{Output:} $\tilde{\mathbf{L}}_S$ with scaled edge weights\\
  \algsetup{indent=1em, linenosize=\small} \algsetup{indent=1em}
    \begin{algorithmic}[1]
    \STATE{Initialize: $k=1$, $\eta^{(1)}=\eta_{max}$, $\Delta_{\lambda_n}=\left(\overline{\Delta}_{\lambda_{n}}\right)^{\frac{1}{N_{max}}}$, $\lambda_1^{(1)}=\lambda_1^{(0)}$, $\lambda^{(1)}_n=\lambda_n^{(0)}$, $\Delta \lambda^{(1)}_1=\lambda_1$}; 
     \STATE{Do initial subgraph edge scaling by $w^{(1)}_{pq}=\frac{w^{(0)}_{pq}\sqrt[]{\lambda_1^{(0)}/\lambda_n^{(0)}}}{10}$ for each edge $(p,q)\in E_S$};
     \WHILE {$\left(\mathbf{\frac{\Delta \lambda^{(k)}_1}{\lambda^{(k)}_1}\geq\epsilon}\right)$~$\wedge$~$\left(k\leq N_{max}\right)$}
     	\STATE{Compute approximate eigenvector $\mathbf{h}^{(k)}_t$ by (\ref{formula_criticality})};
        \FOR{each edge $(p,q)\in E_S$}
        \STATE{$s^{(k)}_{pq}:=-\lambda^{(k)}_1 \left(\mathbf{e}_{pq}^\top\mathbf{h}^{(k)}_t\right)^2$, ~~~$\Delta w^{(k+1)}_{pq}:=\beta \Delta w^{(k)}_{pq}-\eta^{(k)} s^{(k)}_{pq}$};
        \STATE{$\phi(p):=\frac{\mathbf{d}_G(p)}{\mathbf{d}_S(p)+\Delta w^{(k+1)}_{pq}}$, ~~~$\phi(q):=\frac{\mathbf{d}_G(q)}{\mathbf{d}_S(q)+\Delta w^{(k+1)}_{pq}}$};
	\STATE{\textbf{if} $\min\left(\phi(p),\phi(q)\right)\leq \lambda^{(k)}_n \Delta_{\lambda_n}$}
	\STATE{~~~~~$\Delta w_p:=\frac{\mathbf{d}_G(p)}{\Delta_{\lambda_n}}-\mathbf{d}_S(p)$, ~~~$\Delta w_q:=\frac{\mathbf{d}_G(q)}{\Delta_{\lambda_n}}-\mathbf{d}_S(q)$}, ~~~$\Delta w^{(k+1)}_{pq}:=\min\left(\Delta w_p,\Delta w_q\right)$;
        \STATE{\textbf{end~if} }
	\STATE{$w_{pq}:=w_{pq}+\Delta w^{(k+1)}_{pq}$,~~~$\mathbf{d}_S(p):=\mathbf{d}_S(p)+\Delta w^{(k+1)}_{pq}$, ~~~$\mathbf{d}_S(q):=\mathbf{d}_S(q)+\Delta w^{(k+1)}_{pq}$};
        \ENDFOR
        \STATE{$\eta^{(k+1)}:=\frac{\lambda^{(k)}_1}{\lambda_1}\eta_{max}$,~~~$k:=k+1$,~~~and update $\lambda^{(k)}_1$ \& $\lambda^{(k)}_n$};
        \STATE{$\Delta \lambda^{(k)}_1:=\lambda^{(k)}_1-\lambda^{(k-1)}_1$};
     \ENDWHILE
     \STATE {Return $\tilde{\mathbf{L}}_S$}.
    \end{algorithmic}
    }
\end{algorithm}
 To efficiently solve (\ref{edge_opt}),  a constrained SGD algorithm  with momentum \cite{sutskever2013importance} has been proposed for  iteratively scaling up edge weights, as shown in Algorithm \ref{alg:sgd}. The algorithm inputs include: the graph Laplacians $\mathbf{L}_G$ and  $\mathbf{L}_S$,  vectors  $\mathbf{d}_G$ and  $\mathbf{d}_S$  for storing diagonal elements in Laplacians, the largest and smallest generalized eigenvalues $\lambda_1^{(0)}$ and $\lambda_n^{(0)}$ before edge scaling, the upper bound reduction factor $\overline{\Delta}_{\lambda_{n}}$ for $\lambda_n$, the coefficient $\beta$ for combining the previous and the latest updates during each SGD iteration with momentum, the maximum step size $\eta_{max}$ for update, as well as the SGD convergence control parameters $\epsilon$ and $N_{max}$.

The key steps of the algorithm include: \textbf{1)} a random vector  is first generated and used to compute the approximate dominant eigenvector with (\ref{formula_criticality}) as well as edge weight sensitivity with (\ref{formula_weight_sensitivity2}); \textbf{2)}  the   weight update for each edge in the subgraph is estimated based on the previous update (momentum) as well as the latest step size and gradient; \textbf{3)} the impact on the reduction of $\lambda_{n}$ will be evaluated for each weight update to make sure the decreasing rate is not too fast; \textbf{4)} check the cost function $\lambda_{1}$  or the largest weight sensitivity to determine whether   another SGD iteration is needed.

Since edge weights in the subgraph ($\mathbf{w}_{S}$) will be updated during each SGD iteration, we need to solve a new subgraph Laplacian matrix $\mathbf{L}_{S}$ for updating the approximate eigenvector $\mathbf{h}_t$ in (\ref{formula_weight_sensitivity2}), which can be efficiently achieved by leveraging recent sparsified algebraic multigrid (SAMG) algorithm that has shown highly scalable performance for solving large graph Laplacians~\cite{zhiqiang:iccad17}. Since the subgraph topology remains unchanged during the SGD iterations, it will also be  possible to exploit incremental update of graph Laplacian solvers to further improve efficiency.


\section{Filtering Eigenvectors of Sparsified Laplacians}\label{sect:smoothing}
There is a clear analogy between traditional signal processing or classical Fourier analysis and graph signal processing \cite{shuman2013emerging}: 1) the signals at different time points in classical Fourier analysis correspond to the signals at different  nodes in an undirected graph;
2) the more slowly oscillating functions in time domain correspond to the graph Laplacian eigenvectors associated with lower eigenvalues or the more slowly varying (smoother) components across the graph.

Motivated by recent graph signal processing research \cite{shuman2013emerging}, we  introduce a simple yet effective procedure for filtering out errors in Laplacian eigenvectors computed using spectrally sparsified graphs, which  enables to  leverage ultra-sparse subgraphs in SC   for achieving superior solution quality. 

In the following analysis, we  assume that the  $k$ smallest  eigenvalues and their eigenvectors of $\mathbf{L}_{G}$  have been pretty well preserved in $\mathbf{L}_S$ through the proposed two-phase spectral sparsification approach, while the remaining $n-k$ higher eigenvalues and eigenvectors are not. Then the spectral decompositions of $\mathbf{L}_G$ and $\mathbf{L}_S$ of (\ref{formula_decomposition G}) can be written as:
\begin{equation}\label{formula_smoothing_P} 
\begin{array}{l}
\mathbf{L}_G = \sum\limits_{i = 1}^{n-1} {{\zeta _i}{\omega_i}{\omega^\top_i}},\\

\mathbf{L}_{S} =\sum\limits_{i = 1}^{n} {{\tilde \zeta _i}\mathbf{\tilde \omega_i}\mathbf{\tilde \omega^\top _i}}\approx \sum\limits_{i = n-k+1}^{n} {{\zeta _i}\mathbf{\omega_\mathbf{i}}{\omega^\top _i}}+\sum\limits_{i = 1}^{n-k} {{\tilde \zeta _i}\mathbf{\tilde \omega_i}\mathbf{\tilde \omega^\top _i}}. 
\end{array}
\end{equation}

\begin{algorithm}[!htbp]
\small { \caption{Algorithm for Iterative Eigenvector Filtering} \label{alg:jacobi}
\textbf{Input:} $\mathbf{L}_G=\mathbf{D}_G-\mathbf{A}_G$, $\tilde{\omega}_n$,..., $\tilde{\omega}_{n-k+1}$, $\gamma$,$N_{filter}$ \\
\textbf{Output:} The smoothed eigenvectors.\\
  \algsetup{indent=1em, linenosize=\small} \algsetup{indent=1em}
\begin{algorithmic}[1]
    \STATE{For each of the approximate eigenvectors  $\tilde{\omega}_1$,..., $\tilde{\omega}_k$, do}
    \FOR{$i=1$ \textbf{to} ${N_{filter}}$ \textbf}
    \STATE{$\tilde{\omega}^{(i+1)}_{j}=(1-\gamma)\tilde{\omega}^{(i)}_{j}+\gamma\mathbf({{D}_G-{\tilde\zeta_j}})^{-1}\mathbf{A}_G\tilde{\omega}^{(i)}_{j}$}
    \ENDFOR
     \STATE{Return the smoothed eigenvectors $\tilde{\omega}_n$,..., $\tilde{\omega}_{n-k+1}$. }
\end{algorithmic}
}
\end{algorithm}
In the following, we show that using existing sparse eigenvalue decomposition methods for computing the first few Laplacian eigenvectors using spectrally sparsified graphs will  introduce errors expressed as linear combination of eigenvectors corresponding to only large eigenvalues. Since the power iteration method is well known  for calculating a few extreme eigenvalues and eigenvectors, we analyze the error introduced by sparsified Laplacians in power iterations, while the errors by other algorithms can be similarly analyzed in probably more complicated ways. To compute  the smallest eigenvalues and their eigenvectors for  $\mathbf{L}_G$, a few  inverse power iterations can be applied:
\begin{equation}\label{replace power method G}
(\mathbf{L}_{G}+z\mathbf{I})^{l}\mathbf{ x}=\mathbf{r^\perp},
\end{equation}
where $l$ is the number of inverse power iterations, $\mathbf{r^\perp}\in R^n$ is a random vector orthogonal to the all-one vector $\mathbf{1}$, $z$ is a  small positive real number, and $\mathbf{I}\in R^{n\times n}$ is an identity matrix. $z\mathbf{I}^{}$ is added to the Laplacian matrix to make sure that the resultant matrix is non-singular. 

 Let ${\mathbf{x}}$ and $\tilde{\mathbf{x}}$  denote the true and approximate  solutions obtained with $\mathbf{L}_{G}$ and $\mathbf{L}_{S}$, respectively, then we have:
\begin{equation}\label{sol G}
\mathbf{ x}=\sum\limits_{i = 1}^{n} \frac{\omega_i\omega^\top _i\mathbf{r^\perp}}{\left(\zeta _i+z\right)^l};~~~~~~ \tilde{\mathbf{ x}}\approx  \sum\limits_{i = 1}^{n-k} {{{\frac{{\tilde\omega_i}{\tilde \omega^\top _i\mathbf{r^\perp}}}{({\tilde \zeta _i+z})^{l}}}}}+\sum\limits_{i = n-k+1}^{n} {{{\frac{{\omega_i}{ \omega^\top _i\mathbf{r^\perp}}}{({ \zeta _i+z})^{l}}}}},
\end{equation}
which allows us to express the error vector $\mathbf{e}$ as:
\begin{equation}\label{err}
\mathbf{e}=\mathbf{ x}-\tilde{\mathbf{x}} \approx\sum\limits_{i = 1}^{n-k} \left(\frac{{\omega_i}{ \omega^\top _i\mathbf{r^\perp}}}{({ \zeta _i+z})^{l}}- \frac{{\tilde\omega_i}{\tilde \omega^\top _i\mathbf{r^\perp}}}{({\tilde \zeta _i+z})^{l}}\right),
\end{equation}
which shows that when using power iterations to compute the smallest nonzero eigenvalue  and its eigenvector  using sparsified graph Laplacians, the error in the eigenvector can be expressed as a linear combination  of the eigenvectors corresponding to relatively large eigenvalues. If we consider each eigenvector as a signal on the graph, then the eigenvectors of large eigenvalues can be considered as highly oscillating or high frequency signals on graphs. Therefore, the error due to the sparsified graph Laplacian in inverse power iterations can be considered as a combination of high frequency signals on graphs, which thus can be efficiently filtered out using ``low-pass" graph signal filters \cite{shuman2013emerging}. 
In fact, weighted Jacobi or  Gauss-Seidel methods can be efficiently applied for filtering out such high frequency error signals on graphs, which have been widely adopted in modern iterative methods for solving large sparse matrices, such as the smoothing (relaxation) function in multigrid algorithms. This work adopts a weighted Jacobi iteration scheme for filtering eigenvectors on the graph, while the detailed filtering algorithm has been described in Algorithm \ref{alg:jacobi}.

\section{Algorithm Flow and Complexity Analysis}
\begin{figure}
\includegraphics[scale=0.369]{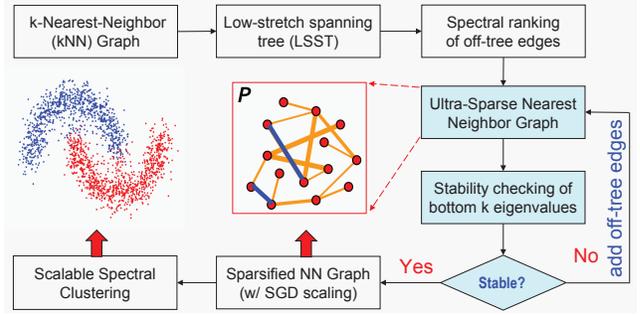}
\caption{The proposed algorithm flow of two-step spectral sparsification for scalable spectral clustering.\protect\label{fig:sc_flow}}
\end{figure}

The proposed SGD-based spectral sparsification method allows computing nearly-linear-sized subgraphs in nearly-linear time, which will significantly accelerate spectral clustering tasks for even very large data sets. In the following, we summarize the key steps of the proposed  method (as shown in Fig. \ref{fig:sc_flow}) as well as their computational complexities:
\begin{enumerate}
   \item Identify the edges to be kept in the subgraph using the spectral graph topology sparsification methods and perform the eigen-stability checking  in \textbf{$O(m\log n)$} time; 
   \item Scale up edge weights in the subgraph using the proposed iterative SGD  scheme (Algorithm \ref{alg:sgd}) in \textbf{$O(m)$} time; 
   \item Perform eigenvalue decomposition for the sparsified graph Laplacian to find the first few eigenvectors. For ultra-sparse Laplacian matrices with tree-like graphs, popular Krylov subspace iterative methods can be highly-scalable in practice due to the nearly-linear time cost for sparse matrix vector multiplications that  dominate the overall computation time;
      \item Apply the proposed eigenvector filtering scheme (Algorithm \ref{alg:jacobi}) to   improve the approximation of Laplacian eigenvectors in \textbf{$O(m)$} time;
\item Run k-means algorithm using approximate eigenvectors for SC in \textbf{$O(knq)$} time \cite{yan2009fast} or \textbf{$O(nq)$} time \cite{kumar2004simple}, where $k$ is the number clusters, and $q$ is the number of iterations.
\end{enumerate}

\vspace{-0.1in}

\section{Experimental Evaluation}\label{sect:experiments}
Extensive experiments have been conducted to demonstrate the effectiveness and efficiency of the proposed method  using the most widely-used real-world benchmark data sets for evaluating the performance of SC algorithms. All experiments for SC have been performed using MATLAB R2018a running on a PC with a $2.50$ GHz Intel Core i5 CPU and $8$ GB RAM. The proposed two-phase spectral graph sparsification algorithm has been implemented in C++.
The reported numbers in our results has been  averaged over $20$ runs.
\vspace{-0.1in}

\subsection{Experiment Setup}
The five real-world data sets used in our experiments can be downloaded from the UCI machine learning repository \footnote{https://archive.ics.uci.edu/ml/} and LibSVM Data\footnote{https://www.csie.ntu.edu.tw/~cjlin/libsvmtools/datasets/}. We briefly describe them as follows: \textbf{COIL-20} includes 1,440 gray scale images of 20 different objects and each image is represented by $1,024$ attributes; \textbf{PenDigits} includes  $7,474$ handwritten digits and each digit is represented by $16$ attributes;
\textbf{USPS} includes   $9,298$ images of USPS hand written digits with $256$ attributes;
\textbf{MNIST} is a data set from Yann LeCun's website\footnote{  http://yann.lecun.com/exdb/mnist/}, which includes  $70,000$ images of hand written digits with each of them represented by $784$ attributes;
\textbf{Covtype} includes  $581,012$ instances for predicting forest cover type  from cartographic variables and each instance with $54$ attributes is from one of seven classes.
 The statistics of these data sets are shown in Table~\ref{table:benchmark}.

\begin{table}
\begin{center}
\normalsize\addtolength{\tabcolsep}{-2.5pt} \centering
\caption{Statistics of the data sets.}
\begin{tabular}{ |c|c|c|c|c|c|  }
 \hline   Data set        &   Size   &  Dimensions    &  Classes \\
 \hline   COIL-20          &  1,440  &  1,024   &    20           \\
 \hline   PenDigits          &  7,494  &  16   &    10           \\
 \hline   USPS     &  9,298  &  256   &    10          \\
 \hline   MNIST          &  70,000  &  784   &    10           \\
 \hline   Covtype         &  581,012  &  54   &    7           \\
 \hline
\end{tabular}\label{table:benchmark}
\end{center}
\end{table}

We compare the SC method  accelerated by the proposed SGD-based spectral sparsification method  against the following state-of-the-art algorithms:
(1) the \textbf{Original SC algorithm} \cite{chen2011parallel},
(2) the \textbf{Nystr\"om method} \cite{fowlkes2004spectral},
(3) the {landmark-based SC method} that uses k-means for landmark selection (\textbf{LSCK}) \cite{chen2011large}, 
(4) the {landmark-based SC method} that uses random sampling for landmark selection (\textbf{LSCR}) \cite{chen2011large}, and
(5) the \textbf{KASP  SC algorithm} using k-means \cite{yan2009fast}.

For fair comparison, we use the same parameter setting in \cite{chen2011large} for compared algorithms: the number of sampled points in \textbf{Nystr\"om method} ( or the number of landmarks in \textbf{LSCK} and \textbf{LSCR}, or the number of centroids in \textbf{KASP} ) is set to 500. We use MATLAB inbuilt kmeans function for all the algorithms.
\vspace{-0.1in}

\subsection{Parameter Selection}
For parameters of our method, to create k-nearest neighbor graphs, $k$ is set to be $10$ for all data sets. We use the self-tuning Gaussian kernel for converting the original distance matrix to the affinity matrix. The off-tree edge budget $b$ measures the amount of off-tree edges added to the LSST for the spectral graph topology sparsification phase, which is defined as $b=\frac{|E_S|-|V|+1}{|V|}$. The number of generalized power iterations  for spectral embedding is set to be $t=2$. The parameters for the  edge scaling  (Algorithm \ref{alg:sgd}) are:
$\overline{\Delta}_{\lambda_{n}}=0.5$, $\beta=0.5$, $\eta_{max}=0.2$, $\epsilon=0.01$, and $N_{max}=100$. The parameters for the  eigenvector filtering (Algorithm \ref{alg:jacobi}) are: $\gamma=0.7$, and $N_{filter}=10$.
The off-tree edge budget $b$ is set to be less than $0.15$ for all data sets.

We assess the clustering quality by comparing the labels of each sample obtained by performing clustering algorithms with the ground-truth labels provided by the data set. The clustering accuracy (ACC)  is used to evaluate the performance, which is defined as $ACC= \frac{\sum\limits_{j = 1}^n  {\delta {(y_i,map(c_i))}}}{{n}}$, where $n$ is the number of data instances in the data set, $y_i$ is the ground-truth label provided by the data set,and $c_i$ is the label generated by the clustering algorithm. $\delta (x,y)$ is a delta function defined as: $\delta (x,y)$=1 for $x=y$, and $\delta (x,y)$=0,  otherwise. $map(\bullet)$ is a permutation function that maps each cluster index $c_i$  to a ground truth label, which can be realized using the Hungarian algorithm \cite{papadimitriou1982combinatorial}. A higher value  of $ACC$ indicates a better  clustering accuracy (quality).

\vspace{-0.1in}

\subsection{Experimental Results}
The runtime and ACC results obtained using different algorithms are provided in Table~\ref{table:clustering time pc}. We also report the time cost of sparsification (Ts) and the speedup of our method compared to the original method in Table~\ref{table:clustering time pc}. The clustering runtime includes the eigenvalue decomposition time and the k-means time. For large data sets such as \textbf{MNIST} and \textbf{Covtype}, our method achieved \textbf{3,571X} and \textbf{9,703X} times speedup, respectively. For the \textbf{Covtype} data set, the proposed method achieves a significantly better  accuracy level than any of the other approximate SC methods. Such a high quality  is mainly due to the guaranteed preservation of key Laplacian eigenvalues and eigenvectors enabled by the proposed method. On the other hand, without robust preservation of the original graph spectrum,  the  SC results  by existing approximate SC methods can be quite misleading. For example, the KASP, LSCK and LSCR algorithms that use simple sampling methods, such as k-means and random sampling, may lead to substantial loss of  structural information of the data set, and thus poor accuracy in SC tasks; the approximation quality of the Nystr\"om method strongly depends on the encoding power of the data points chosen from k-means clustering \cite{choromanska2013fast}, but for large data set it is very unlikely that the small amount chosen data points can truthfully encode the entire data set.  We also observe that the  ACC results obtained by using spectrally sparsified graphs are even better than the results obtained by using the original graphs for most data sets, indicating that the proposed two-phase spectral sparsification method  does help remove spurious edges, leading to improved graph structures for SC tasks.

\begin{table*}
\begin{center}
\scriptsize\addtolength{\tabcolsep}{-2.5pt}\centering
\caption{Clustering accuracy (\%) and clustering (eigendecomposition and k-means) time (seconds)}  
\begin{tabular}{ |c|c|c|c|c|c|c|c|c|c|c|c|c|c|c|}
\hline
 &\multicolumn{6}{|c|}{Clustering Accuracy (ACC)} &\multicolumn{7}{|c|}{Spectral Clustering Time}\\
 \hline Data Set&Original& Nystr\"om & KASP &LSCK&LSCR&Ours&Original&Nystr\"om &KASP &LSCK&LSCR &Ours&Ts\\
 \hline  COIL-20   &78.80  &67.44 &58.83 & 72.41&68.45&\textbf{76.27}&0.37&0.46&2.74 &2.44&0.23&0.18 (2.05X)&0.1\\
 \hline  PenDigits  &81.12 &68.70 &75.83 &80.77&77.89  &\textbf{83.26}&0.47&0.28 &1.00  &0.81& 0.23&0.26 (1.80X)&0.1\\
 \hline USPS  & 68.22  &68.83 &72.61 &77.54&66.22  &\textbf{70.74} &1.02 &0.40  &6.88  &7.08&0.24&0.20 (5.10X)&0.1\\    
 \hline  MNIST  &71.95   &53.27 & 68.03 &69.88&57.24 &\textbf{72.27}&6785 &0.80 &754 &722 &0.81&1.90 (\textbf{3,571X})&3.5\\
 \hline Covtype &48.83  &24.78  &27.11  & 22.80&22.79&\textbf{48.86}&91,504&18.51 &1,165 &1,154 &7.23&9.43 (\textbf{9,703X})&10.9\\
 
 \hline
\end{tabular}\label{table:clustering time pc}
\end{center}
\end{table*}

 Fig.~\ref{fig:eig_stable} shows the impact of adding off-tree edges to the stability (variation ratio) of the bottom eigenvalues computed by (\ref{eqn:varaition ratio}),which verifies the theoretical foundation of the proposed method. Adding extra off-tree edges will immediately reduce the variation ratio of the bottom eigenvalues,  indicating a gradually improved eigenvalue stability.

We want to note that a dramatically sparsified graph not only leads to more efficient eigenvalue decomposition procedure, but also significantly improved memory efficiency in data clustering. Since most clustering algorithms are in-memory algorithms \cite{li2016scalable}, all data and intermediate results have to be stored in the main memory to achieve good computing efficiency. Consequently, the tree-like ultra-sparse graphs computed by the proposed method will allow more scalable  performance especially when dealing with very large  data sets.

\begin{figure*}
\begin{center}
\includegraphics[scale=0.2501999]{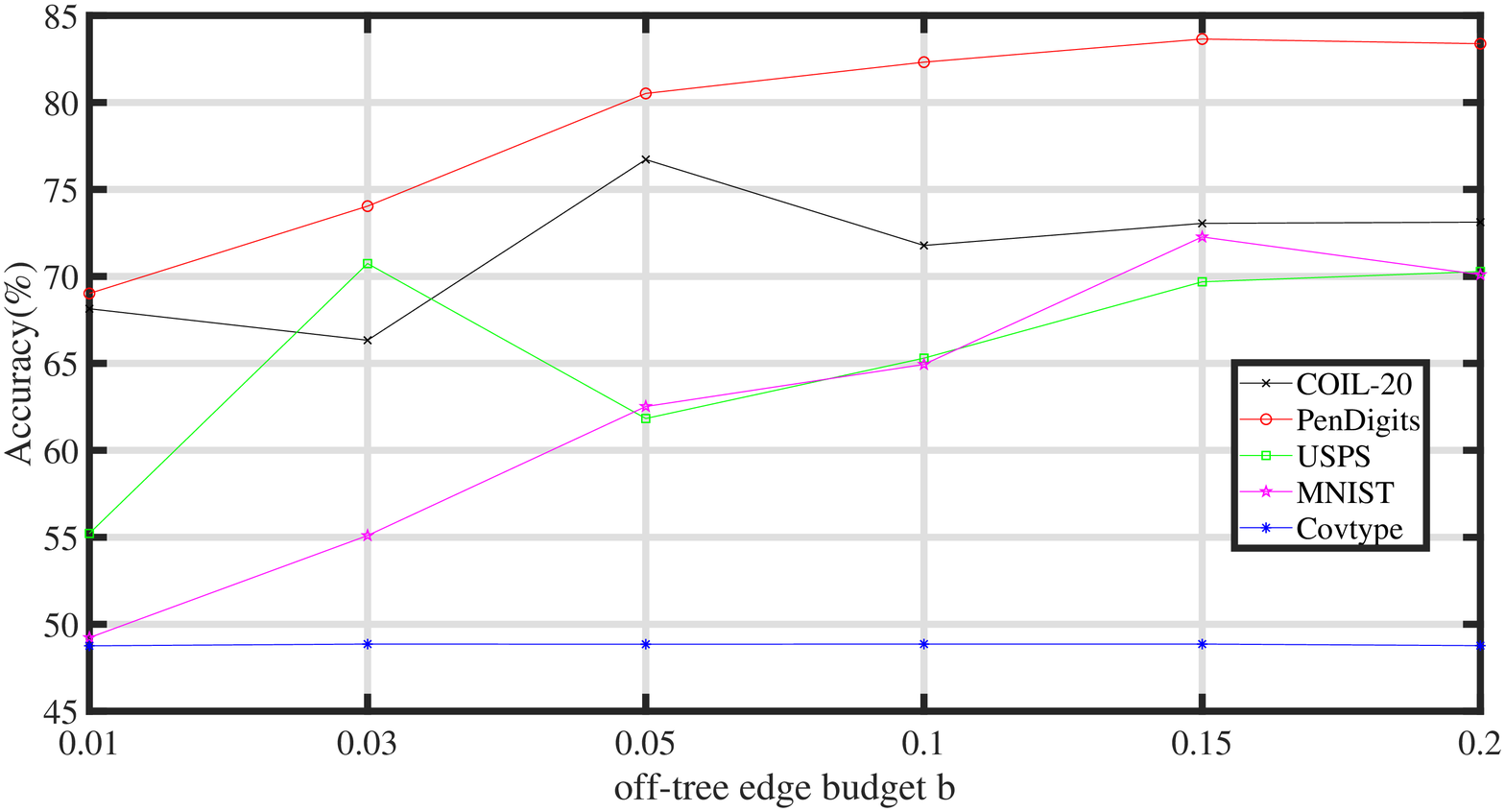}\includegraphics[scale=0.2501999]{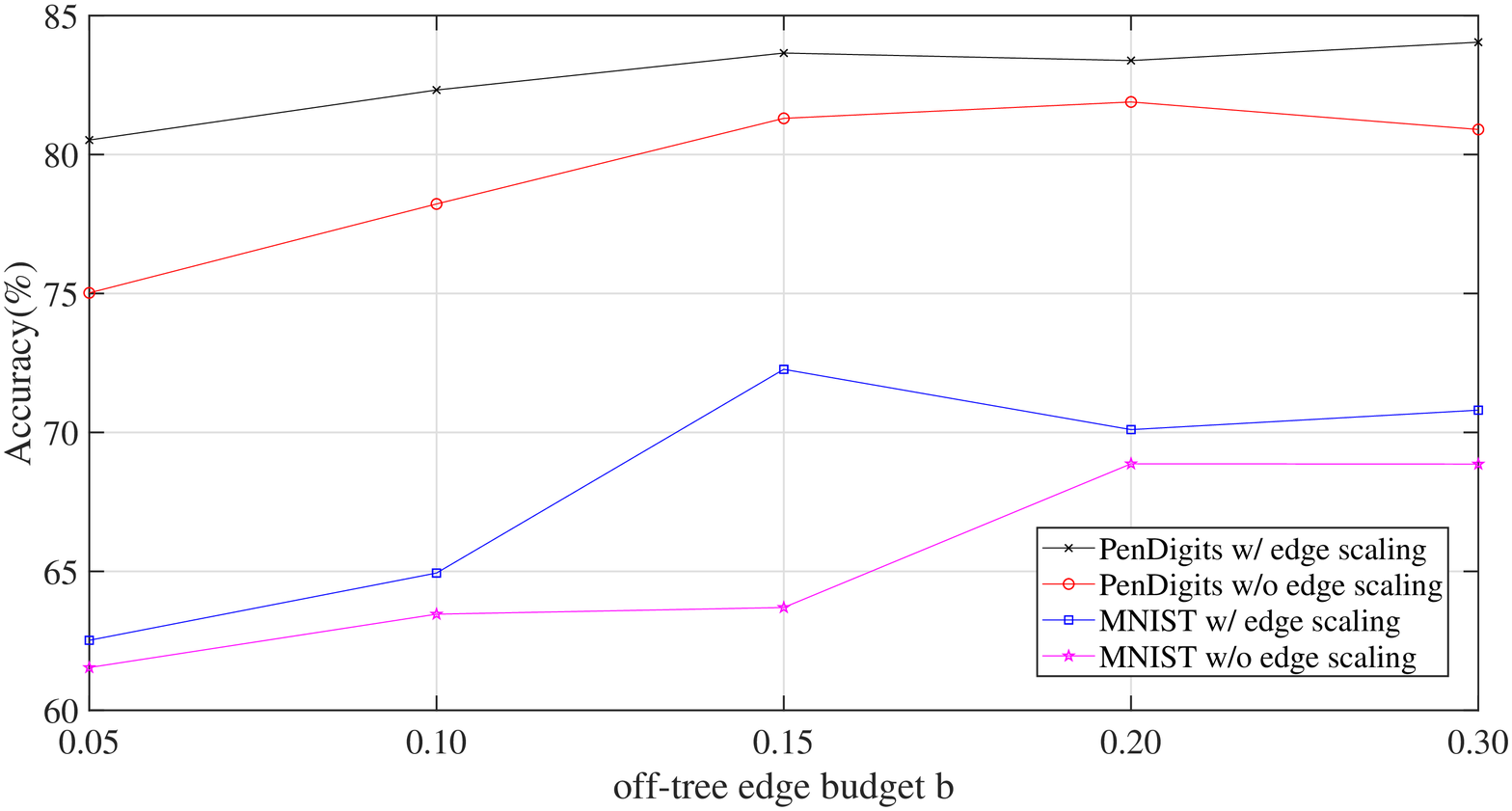}
\caption{Left: ACC VS off-tree edge budget b; Right: ACC w/ and w/o edge scaling \protect\label{fig:accscaling}}
\end{center}
\end{figure*}


The ACC results with respect to different options of off-tree edge budget ($b$)  have been shown in Figure~\ref{fig:accscaling}. As observed, adding only $0.01|V|$ of off-tree edges will suffice for achieving the peak ACC result for the \textbf{Covtype} data set. For  other  data sets, the ACC results gradually approach the best ones after adding less than $0.15|V|$ of off-tree edges to the spanning tree.
We also show the effectiveness of the proposed  edge scaling phase in the figure. As observed, even with much greater number of off-tree edges added into the LSST, the sparsified graphs without edge scaling  still can not reach the  clustering accuracy level achieved by the subgraphs obtained with edge scaling.   
\section{Conclusions}\label{sect:conclusions}
To fundamentally address the computational challenge due to the eigenvalue decomposition procedure in SC, this work introduces a two-phase spectral sparsification framework that enables to construct tree-like subgraphs with guaranteed preservation of original graph spectra. Our method includes a nearly-linear time spectral graph topology sparsification phase and an iterative edge weight scaling phase.  Our results on a variety of public domain data sets show  dramatically improved clustering performance when compared with state-of-the-art  SC methods.

\end{document}